# SINGLE IMAGE SUPER RESOLUTION IN SPATIAL AND WAVELET DOMAIN

Sapan Naik[1], Nikunj Patel[2]

[1]Department of Computer Science and Technology,
Uka Tarsadia University, Bardoli, Surat, India
Sapan_say@yahoo.co.in
[2]Department of Computer Engineering,
Sardar Vallabhbhai Institute of Technology, Vasad
Nikunj_patel10@yahoo.in

## ABSTRACT

*Recently single image super resolution is very important research area to generate high-resolution image from given low-resolution image. Algorithms of single image resolution are mainly based on wavelet domain and spatial domain. Filter's support to model the regularity of natural images is exploited in wavelet domain while edges of images get sharp during up sampling in spatial domain. Here single image super resolution algorithm is presented which based on both spatial and wavelet domain and take the advantage of both. Algorithm is iterative and use back projection to minimize reconstruction error. Wavelet based denoising method is also introduced to remove noise.*

## 1. INTRODUCTION

High-resolution images are requisite in solicitations like high quality videoconference, remote reconnaissance, high definition television broadcasting, and medical imaging. Due to precincts like camera cost, power, memory size and limited bandwidth, it is not always possible to get high-resolution image. To acquire high-resolution image from low-resolution image in such solicitations, algorithms are requisite. Basic tenacity of super resolution algorithm is to create high-resolution image from low-resolution image, which looks as if the image is captured from high-resolution camera. Preliminary Super resolution algorithms work on multiple images of same sight and generate one image [19]. In recent time, algorithms for real time and single frame super resolution have been evolved and developed.

Some of the most recent work on single image super resolution has been done using texture hallucination, patch based up sampling and example based super resolution [3,14]. In edge-based algorithms, some edge priors are used to reconstruct sharp images but problem with these methods are they produce blurriness and over smoothness in some regions [8]. Freeman et al. proposed example-based super-resolution approach [12] in which patch-based image model based on training database is used. Method generates good result but it not produces consistent texture and generates some noise also.

The relation between the high- resolution and low-resolution images is the reason that has accelerated us to the popular study and discussion on Spatial-domain based interpolation.

Interpolation techniques like pixel replication and bilinear interpolation up sample an image without considering any details of input image. These methods work well in smooth region but edges and some textures get blurred. In wavelet-domain based techniques of image interpolation the foremost challenge is to estimate unknown coefficients of three high frequency sub bands. Details of wavelet are given in the later section. Basic interpolation method in wavelet domain is Wavelet Zero-Padding. In this method low resolution image is multiply with scaling factor S, which work as top left quadrant (LL) of final high resolution image. In other three quadrants of high-resolution image (HH, HL and HH), zeros are padded. Temizel [1] combines directional cycle spinning method with Wavelet Zero-Padding (WZP) interpolation method.

In this paper single image super resolution algorithm is proposed which uses both spatial and wavelet domain. For up sampling and down sampling of an image in spatial domain, respectively bicubic smoother and bicubic sharper method of adobe Photoshop cs5 is used. For removing blur and get smoother result back projection method is used. Wavelet based denoising method is also used to remove noise from image.

Paper is systematized as follows: section 2 contains proposed algorithm based on wavelet as well as spatial domain. Section 3 spectacles experiments details and results. Finally section 4 comprises conclusion and future work. Sample MATLAB code is also provided.

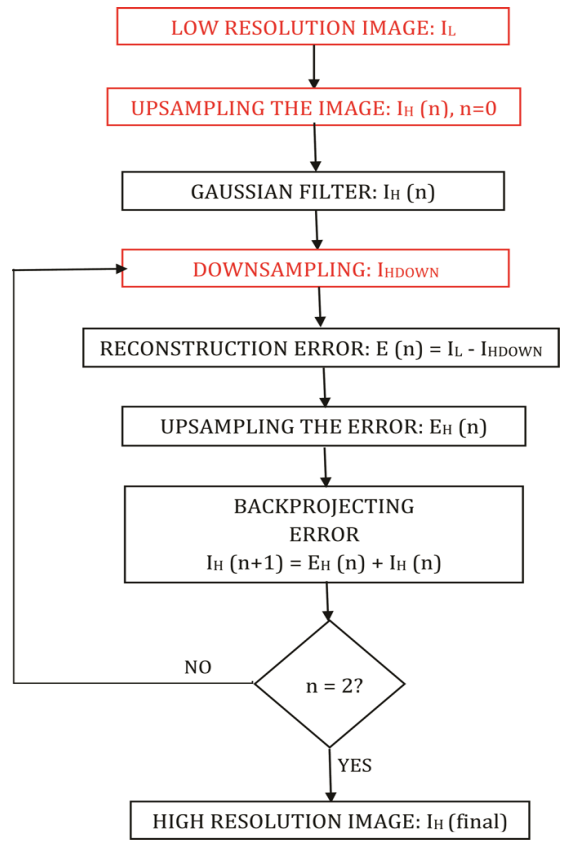

Figure 1.  Proposed Algorithm

## 2. PROPOSED ALGORITHM

Figure 1 displays the flowchart of proposed algorithm for single image super resolution that is based on the algorithm presented in [11]. Here some changes have been done in original algorithm of [11] as down sampling method is changed, new up sampling algorithm is used and wavelet based denoising algorithm is introduced. Below each step of flowchart is explained in detail.

## 2.1. Step 1: Down sampling to get low-resolution image

To acquire the low-resolution image, take high-resolution image and convert it into low-resolution image using below method (Figure 2). Here block of 2 x 2 is chosen and 1 pixel of low-resolution image for every 4 pixel of high-resolution image is derived.

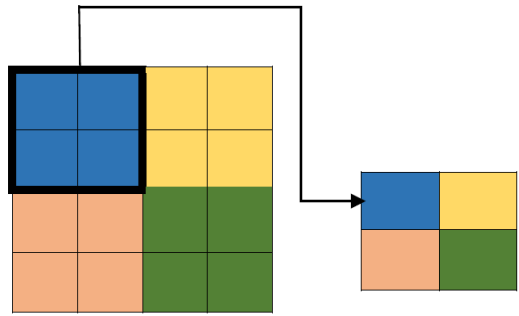

Figure 2. Method of down sampling

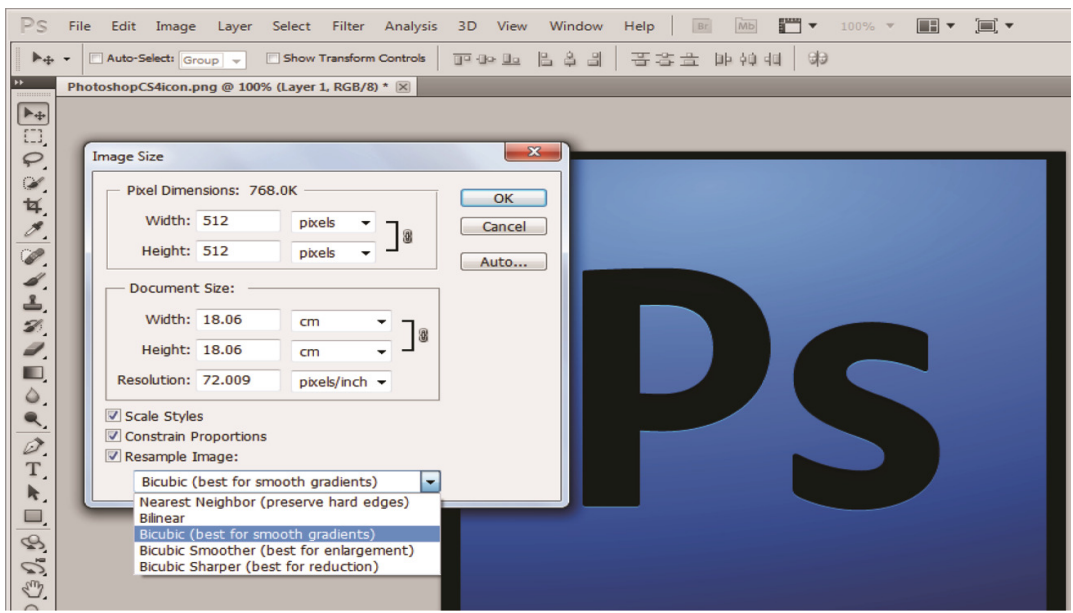

Figure 3. Snapshot of Photoshop showing different interpolation methods

As shown in the figure 3, in order to acquire low resolution image, directly Bicubic sharp option of Photoshop cs5 can be chosen, the reason is that Bicubic Smoother is a good method for enlarging images that are based on Bicubic interpolation but designed to produce smoother results moreover it is also used for reducing the size of an image based on Bicubic interpolation with enhanced sharpening and maintains the detail in a resampled image [4]. In this paper, for experiment purpose this method has been used.

## 2.2. Step 2: Up sampling of the image

Method proposed in [9] is used for up sampling. Some of the changes have been made in original method like instead of Bicubic, Bicubic smoother method of Photoshop cs5 (as shown in figure) is used, denoising algorithm is applied on HH band and HAAR wavelet is chosen and instead of Cohen-Daubechies-Feauveau (CDF) 9/7 wavelet as a mother wavelet transform because HAAR wavelet is computationally fast. Many other wavelets are available which provide better results like sym; db4 etc. but all require more time for computation. During experiment on program using HAAR, it has taken 8.37 seconds while DB4 has taken 9.13 seconds. New modified algorithm for image up sampling is shown in the figure. Below explained is the complete method of image up-sampling.

As shown in figure 4, apply SWT on low-resolution image of size m x n which produce four sub bands (ll, lh, hl, hh) with size of m x n each. SWT is same as Discrete Wavelet Transform (DWT) but SWT generates each sub band of the size of image while in DWT each sub band is half the size of image.

For signal decomposition, one can use analysis filter bank which consist of low pass and high pass filters at each decomposition stage and split signal into two bands. The low pass filter fetch the coarse information (corresponds to an averaging operation) while high pass filter fetch detail information (corresponds to a differencing operation) of the signal. Finally divide the output of filtering operations' by two [10,15,16].

For two-dimensional transform, the image is filtered along the x-dimension using low pass and high pass analysis filters and decimated by two. Then it is followed by filtering the sub-image along the y-dimension and decimated by two. Finally, the image has been split into four bands denoted by LL, HL, LH, and HH, after one level of decomposition [17,18]. The LL band is again subject to the same procedure. This process of filtering the image is called pyramidal decomposition of image. This is depicted in Fig. Reversing the above procedure can carry out the reconstruction of the image and it is repeated until the image is fully reconstructed [10].

Apply Bicubic smoother interpolation of Photoshop cs5 as expounded in step1 on low-resolution image that produces up-sampled image. On these up-sampled image, apply DWT which produces four sub bands of size m x n each. Now LH, HL and HH sub bands produced by DWT and by SWT are incremented to correct the estimated coefficients.

Now apply denoising algorithm on HH sub band only, because LL sub band contains main information about the image while main noise is present in other three sub bands and maximum high frequency noise present in HH. Below given is the description of denoising algorithm that is based on the algorithm presented in [10].

To remove additive noise and maintain the important details of the image at the same time Denoising techniques are essential. DWT based denoising method gives good result as wavelet transform contains large coefficients of images, which represents the detail of image at different resolutions. Two methods are available for denoising, hard Thresolding and Soft Thresolding [2].

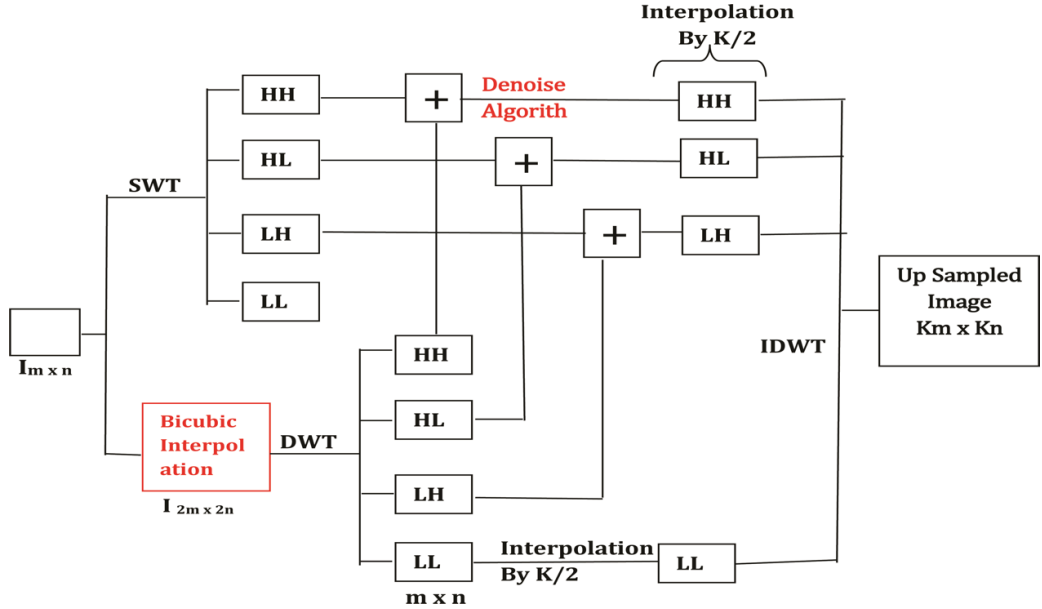

Figure 4. Proposed image up sampling method

Hard Thresholding

$$I(P,T) = P \text{ if } |P| > T \text{ and } I(P,T) = 0 \text{ if } |P| < T \qquad (1)$$

Soft Thresholding

$$I(P,T) = sign(P) * max(0, |P| - T) \qquad (2)$$

Where T is the threshold level, P is the input sub band and D is the denoised band. Algorithm used the Median Absolute Deviation (MAD) to calculate noise level.

$$\sigma = median\ |S_{i,j}|\ /0.6745 \qquad (3)$$

Where $S_{i,j}$ = LH, HL, HH and Threshold value is calculated by

$$T = \sigma - (|\text{Harmonic Mean} - \text{Geometric Mean}|) \qquad (4)$$
$$\text{Where Harmonic Mean} = M2/ \sum_{i=1}^{M}\sum_{j=1}^{M} 1/g(i,j) \qquad (5)$$
$$\text{And Geometric Mean} = [\prod_{i=1}^{M}\prod_{j=1}^{M} g(i,j)]^{(1/M2)} \qquad (6)$$

So the procedure is, from HH sub band calculate noise level(σ) than find threshold value(T) for that sub band and finally apply soft thresolding method to get denoise HH sub band. Finally interpolate all four-sub bands using bicubic smoother with K/2 factor and apply inverse DWT to get up sampled image with the km x kn size.

## 2.3. Step 3 and 4: Gaussian Filter and Down sampling

After up sampling due to point spread function (PSF) image can be look blurred little bit. So Gaussian filter merely work like smoothing kernel. As the blurred effect is very low or ignorable Gaussian filter applies only once. Instead of Gaussian filter, winey filter can be used or Iterative blur DE convolution, Lusy Richardson algorithm can be used[6]. Same as step1 using bicubic sharper algorithm image is down-sampled.

### 2.4. Step 5 and 6: Reconstruction and up sampling the error

In this stage of algorithm error is calculated between original low-resolution image of step 1 and down sampled image of step 4. This is the most important part of algorithm because the error that we find in this stage is used as the correction parameter in getting super resolution image and also used for refining coefficients of sub bands. By experiments it has been observed that after three to four iteration error becomes so small that it can be neglected.

Up sampling the error is most important step of the proposed algorithm. For reconstructing super resolution image, error must be back projected and for that error matrix must be up-sampled to meet super resolution image. Bicubic smoother algorithm of Photoshop cs5 is used for up-sampling error matrix.

### 2.5. Step 7: Back projecting error

Finally error matrix generated in step 6 is added with high-resolution image generated in step 3. Repeat the above procedure as shown in the figure till we acquire satisfactory results. Within three iterations appropriate result comes.

## 3. EXPERIMENT AND RESULT

Experiment of proposed algorithm is performed on the computer with configuration of Intel i3 processor, 4GB RAM and 512MB NVidia graphics card. For performance evaluation of algorithm, PNSR ratio and visual quality are considered as parameters. The PSNR is defined as: PSNR = 20 · log 10(MAX i / √MSE). Comparison of cubic interpolation, new edge directed interpolation [33], wavelet zero padding (WZP) same as image up-sampling using DWT [55], algorithm proposed in [11](denoted as FF) and proposed method have been done for six different images of lina, Mona Lisa, Baboon, Pepper, Dog, PS logo and some textures (A1 to A6). For doing up sampling in algorithm, Photoshop is used. Same algorithm is also implemented in MATLAB without using Photoshop (used bicubic interpolation function). It gives almost same result as manual method. Sample MATLAB code for proposed algorithm is also given.

Table 1. Comparison of different methods using PSNR ration(512 x 512 resolution)

|           | **Bicubic** | **NEDI** | **WZP** | **FF**  | **Proposed Method** |
|-----------|-------------|----------|---------|---------|---------------------|
| Lena      | 32.549      | 31.479   | 31.725  | 32.586  | 32.613              |
| Mona Lisa | 29.428      | 27.917   | 27.944  | 29.624  | 29.589              |
| Baboon    | 32.842      | 31.619   | 31.925  | 32.862  | 32.933              |
| PS logo   | 26.172      | 25.688   | 26.018  | 26.288  | 26.216              |
| Dog       | 26.656      | 25.438   | 25.912  | 26.731  | 26.693              |
| Peppers   | 28.771      | 27.533   | 27.881  | 28.711  | 28.787              |
| A1        | 30.592      | 29.838   | 30.189  | 31.472  | 31.639              |
| A2        | 29.852      | 29.337   | 29.762  | 30.278  | 30.372              |
| A3        | 31.625      | 30.363   | 30.625  | 31.752  | 31.966              |
| A4        | 30.992      | 30.673   | 30.936  | 31.173  | 31.386              |
| A5        | 31.750      | 30.753   | 30.826  | 32.037  | 32.139              |
| A6        | 32.283      | 31.161   | 31.378  | 32.420  | 32.639              |

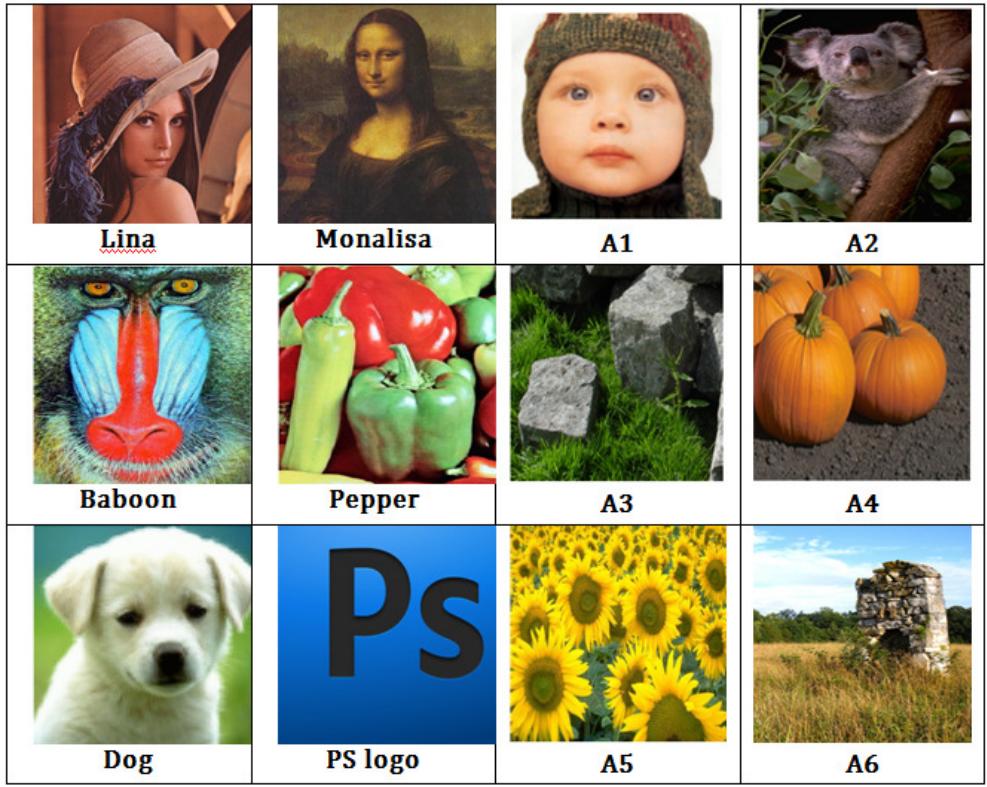

Figure 5.  Input Test Images

## 4. CONCLUSIONS AND FUTURE WORK

Proposed algorithm uses advantage of both wavelet and spatial domain. PSNR ratio and visual quality of images are also shows the effectiveness of algorithm. Algorithm gives almost same result as algorithm proposed in [11]. Proposed algorithm is faster. Some more work on up sampling algorithm will surly improves result. In future more work on wavelet domain and texture-based up sampling will be conducted and comparison will be done to see which algorithm work better for which kind of image.

## SAMPLE CODE

```
%===========step-1===================

x = imread('lena.jpg');
J = imresize(x, 2,'bicubic');
imshow(J);
imwrite(J,'lena1.jpg','jpg')

%===========step-2===================

I=imread('lena.jpg');
[ll1,lh1,hl1,hh1]=swt2(I,1,'haar');
Ib=imread('lena1.jpg');
[ll2,lh2,hl2,hh2]=dwt2(Ib,'haar');
for i=1:256
    for j=1:256
```

```
    lh3(i,j,:)=lh1(i,j,:)+lh2(i,j,:);
    hl3(i,j,:)=hl1(i,j,:)+hl2(i,j,:);
    hh3(i,j,:)=hh1(i,j,:)+hh2(i,j,:);
  end
end

% apply denoising equations

x=idwt2(ll2,lh3,hl3,hh3,'haar');
x = uint8(x);
imshow(x)
imwrite(x,'lena2.jpg','jpg')
```

%=============step-3=================

```
x=imread('lena2.jpg')
bt = 1;
o = 15;
n = 10;
h = gaussfir(bt,n,o);
imf=imfilter(x,h);
imshow(imf);
imwrite(imf,'lena3.jpg','jpg')
```

%=============step-4=================

```
imf = imread('lena3.jpg');
J = imresize(imf, .5);
imshow(J);
imwrite(J,'lena4.jpg','jpg')
```

%=============step-5=================

```
Il=imread('lena.jpg');
J=imread('lena4.jpg');
for i=1:256
   for j=1:256
        E(i,j,:)=Il(i,j,:)-J(i,j,:);
   end
end
imshow(E)
imwrite(E,'lena5.jpg','jpg');
```

%=============step-6=================

```
E = imread('lena5.jpg');
J = imresize(E, 2);
imshow(J);
imwrite(J,'lena6.jpg','jpg')
```

%=============step-7=================
```
Il=imread('lena3.jpg');
J=imread('lena6.jpg');
```

```
for i=1:512
   for j=1:512
         Z(i,j,:)=Il(i,j,:)+J(i,j,:);
   end
end
Z = uint8(Z);
imshow(Z);
imwrite(Z,'lena7.jpg','jpg');
```

%======= Go again to step 4 and do iteration till satisfactory results ========

**Author**

Sapan Naik received his Bachelor degree in 2007 from Dharmsinh Desai University, India. He received his master degree in field of Computer engineering in 2012 from Gujarat Technological University. He is working as Asst. professor at Department of Computer Science and Technology, Uka Tarsadia University, Bardoli, India. His research interest includes Image Processing and machine learning. He has published eight papers at national and international level. He is currently pursuing his Ph.D. on "Smart Farming by Image Processing and Machine Learning" at Uka Tarsadia University

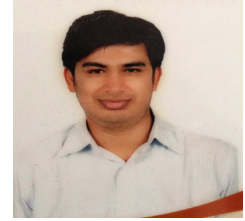